# Register Variation Remains Stable Across 60 Languages

Haipeng Li
University of Canterbury

haipeng.li@canterbury.ac.nz

Jonathan Dunn
University of Canterbury

jonathan.dunn@canterbury.ac.nz

Andrea Nini
University of Manchester

andrea.nini@manchester.ac.uk

*Abstract*

This paper measures the stability of cross-linguistic register variation. A *register* is a variety of a language that is associated with extra-linguistic context. The relationship between a register and its context is functional: the linguistic features that make up a register are motivated by the needs and constraints of the communicative situation. This view hypothesizes that register should be universal, so that we expect a stable relationship between the extra-linguistic context that defines a register and the sets of linguistic features which the register contains. In this paper, the universality and robustness of register variation is tested by comparing variation within vs. between register-specific corpora in 60 languages using corpora produced in comparable communicative situations: tweets and Wikipedia articles. Our findings confirm the prediction that register variation is, in fact, universal.

**1. Comparing Register Variation Across Languages**

A *variety* of a language is a combination of linguistic features that co-vary together: for example, past tenses and third person pronouns, nouns and determiners. A *register* can be defined as a variety of a language that is associated with a specific context of production (Biber and Conrad 2009). In this way, registers contrast with other types of varieties, such as *dialects* or *sociolects*, which are instead associated with social factors. The relationship between a register and its context is functional in nature: for example, the features of a particular register are used because they respond to the constraints and needs of that situation. For example, the past tense and third person pronouns are tools we need to construct a narrative and their usage therefore correlates with situations in which one of the purposes is to narrate (e.g. a fictional novel but also a biography). In the same way, nominalisations and the passive voice can be useful to remove agency from a text, thus being quite useful in scientific and academic prose.

This deep connection between a register and its context means that both need to be described in order to carry out a register analysis. The language of the register is described by referring to linguistic features, which tend to be lexicogrammatical items. And the context of production tends to be described through an analysis of its contextual configuration, for example using *Situational Parameters* (Biber 1994; Biber and Conrad 2009), a taxonomy of those aspects of an extra-linguistic context that are known to influence language use. For example, these situational parameters describe distinctions between written and spoken usage, the relationship between addresser and addressee, and the purpose of the text.

We begin by briefly defining some key terms that will be used throughout this paper. First, *context of production* and *communicative situation* refer to the non-linguistic attributes of the environment in which a corpus was created. *Register*, then, is a specific set of linguistic features which are associated with and thus predictive of a particular context of production; this a latent variable in the sense that it is not directly observed. Following from this, a *register-specific corpus* is a collection of language data which provides a sample of a given register; these are the actual observations available. Within a corpus, *features* are a vector of observable attributes that represent linguistic properties of that corpus in the aggregate; each individual feature in this sense may not itself be linguistically interpretable out of context. However, taken in the aggregate as a feature set these do represent the linguistic properties of a corpus. From this perspective, a specific set of linguistic features is what characterizes a distinct variety of texts which, in turn, is associated with situational characteristics.

Given this terminology, the goal of this paper is to find out whether the context of production is in fact driving register variation in a cross-linguistic setting. We systematically collect comparable register-specific corpora across 60 languages. Our hypothesis is that the same similarity relationships between these register-specific corpora should exist across languages. The reasoning is that the same context of production is shared across languages. Thus, the register-specific corpora, if register is the main source of variation, should have the same configuration across languages.

Despite some comparative work on registers across languages (most notably Biber 1995), there has not been any systematic assessment of cross-linguistic register variation. Such work is especially important because it tests the extent to which this fundamental notion applies to other languages and cultures beyond English and Indo-European languages in Western contexts.

The unspoken hypothesis behind studies of register variation is that this functional connection between language and context is universal. Thus, there should not be any languages in which this connection is absent. Further, functional linguistics would argue that this context-based variation should be stronger than other sources of variation (such as dialects and sociolects). In other words, given the frequency of lexicogrammatical features, register accounts for more of the observed variation than other dimensions like dialect (Dunn 2021). As a hypothetical example, imagine that two individuals each produce language in two different contexts, as schematized in Table 1. Research in authorship analysis suggests that the differences across registers (A-B and C-D) will be greater than the differences across speakers (A-C and B-D). In other words, register variation is so pervasive that any two samples written by two different individuals in the same situation are typically more similar than any two samples produced by the same individual in different situations.

*Table 1. Production by Context Across Individuals*

|  | **Situation 1:** *Business Email* | **Situation 2:** *Conversation with Friends* |
| --- | --- | --- |
| *Person 1* | Linguistic features A | Linguistic features B |
| *Person 2* | Linguistic features C | Linguistic features D |

Our knowledge of the pervasiveness of register variation remains, however, largely confined to English and closely related languages. How stable or robust does this connection between (i) identifiable varieties of a language and (ii) the extra-linguistic context remain in a highly cross-linguistic setting? If we again imagine the comparison in Table 1 we can substitute *Person 1* with *Language 1* and *Person 2* with *Language 2.* This cross-linguistic comparison is the unanswered question that is addressed in this paper: are the similarities between specific communicative situations stable or comparable across languages from different families with very different grammars that represent different cultures and populations?

This question is operationalized in two ways: first, if the context of production is the primary influence shaping a register, then corpora representing a narrow communicative situation should be more homogeneous. In other words, sub-corpora from a register like Wikipedia should be more similar to one another than sub-corpora from a macro-register like web pages because the former were produced in more similar contexts. Second, corpora that represent a fixed communicative situation (here, Twitter and Wikipedia) should have the same linguistic distance across languages. In other words, because these corpora are drawn from the same context of production their linguistic features should have the same pattern across languages even though those features themselves are language-specific.

It is important to note that, in this comparison across languages, the specific linguistic features in question are unique to each language (Biber 1995). For example, if English is Language 1 then imperative clauses and WH-question constructions would be part of the features of a register. But, if Language 2 were Farsi or Arabic, those grammatical features would not be relevant. Despite these linguistic differences, the hypothesis can still be tested by developing a feature space that is standardized within languages and thus comparable across languages, as discussed further in Section 3. Our basic approach is to use corpus similarity measures which can be validated on their predictions about the boundary between register-specific corpora to create a network of relationships between corpora for each of the 60 languages considered.

After reviewing previous work on register and corpus similarity in Section 2, we present the data and experimental methods in Section 3. The main experiments are presented in Section 4: first, measuring the homogeneity of register-specific corpora across languages; second, measuring relationships between register-specific corpora across languages. Finally, in Section 5 we discuss the implications of these experiments for the construct of register within linguistic theory.

## 2. Related Work

The experiments in this paper rely on the ability to measure corpus similarity across many languages (Li and Dunn 2022). A corpus similarity measure provides a scalar representation of the relationship between two samples from a corpus. Generalizations like homogeneity are estimated from the pairwise similarity of many different subsets of the same corpus. In other words, a corpus of a million words can be divided into 100 chunks of approximately 10k words each. There are hundreds of pairwise relationships between these chunks; taken together, this provides an estimate of corpus similarity that is drawn from many individual observations. When applied to a single corpus, this measure represents homogeneity. When applied to two corpora, this measure represents similarity. The underlying corpus similarity measure used here is a frequency-based statistical measure (Kilgarriff 2001; Fothergill et al. 2016; Dunn 2021; Li and Dunn 2022). A continuous measure, as opposed to a text classifier, supports continuous relationships between communicative situations (Biber et al. 2020).

A frequency-based approach to corpus similarity first constructs a feature space that contains the $k$ most frequent features, where each feature is a word n-gram or character n-gram. The most frequent features from this perspective always represent grammatical information. Each sample is represented by a vector of frequencies, one for each feature. These vectors are then compared using a measure like Spearman's *rho*, where a higher value represents more similar corpora and a lower value represents less similar corpora. Here we review work on frequency-based corpus similarity measures and then contrast this with approaches based on multi-dimensional analysis.

One line of related work uses corpus similarity measures to find geographic variation as opposed to register variation. For example, recent work has constructed web corpora from national top-level domains ( *.ca*, *.uk*) that correspond to countries in which English is widely spoken (Cook and Brinton 2017; Cook and Hirst 2012). With a goal of comparing web corpora to conventionally constructed corpora, this work uses measures based on the chi-square test, spelling variants, and the frequencies of words known to be marked in particular varieties to show that web corpora do, in fact, reflect the corresponding variety of English. Frequency-based corpus similarity measures have also been used to show a consistent agreement between digital sources (the web and tweets) across 9 languages and 84 language varieties (Dunn 2021). In this work, word unigram frequencies and character trigram frequencies are used to calculate Spearman's *rho*; the resulting similarity values are used to evaluate the fluctuation within and between registers for different language varieties. The overall finding is that both geo-referenced web data and social media data are representing the same underlying varieties.

The method used here for measuring corpus similarity is based on these same word n-gram or character n-gram frequency vectors. However, while previous work has focused on only a few languages, this study investigates comparable corpora drawn from 60 languages. Because these languages have not been previously studied in this context, we validate the corpus similarity measure for each language using an accuracy metric (cf., Appendix 1). This accuracy metric is calculated using predictions based on a threshold value for determining whether two samples come from the same or different register-specific corpora (cf., Kilgarriff 2001). This larger family of methods, validated on its ability to make distinctions between register-specific corpora, is similar to discriminant analysis (cf., Egbert and Biber 2018).

Within studies of register variation, an alternate method is based on factor analysis, such as multi-dimensional analysis (Biber 1988). The advantage of such dimension reduction methods is that they provide bundles or dimensions of features to describe the functional differences between registers. An approach based on corpus similarity, on the other hand, builds clusters of related samples rather than bundles of related features. Thus, as explored in more detail in Section 3.3, corpus similarity measures can be used to find relationships within subsets of a corpus in a way that reveals functional groupings. From a practical perspective, one approach to situating new corpora within studies of register variation is to apply the Multidimensional Analysis Tagger (Nini 2019), which computes the loadings for an English text for each of the six major dimensions of register-specific features in English. However, a tagger like this requires annotated training data as well as extensive benchmarking corpora. From a cross-linguistic perspective, the specific bundles of features (factor analysis) will be language-specific but the clusters of text types (corpus similarity) should be universal. For these reasons we use corpus similarity as a more scalable and cross-linguistic representation of register variation.

Previous work has used multi-dimensional analysis on a large Brazilian Portuguese corpus representing spoken and written registers (Sardinha, Kauffmann, and Acunzo 2014), the Brazilian Register Variation Corpus (CBVR; Corpus Brasileiro de Variação de Registro). This corpus contains 48 different registers, 12 spoken and 36 written. Six dimensions of linguistic variation are identified using multi-dimensional analysis. Other recent work has used multi-dimensional analysis to investigate structural and functional variation among different English web registers (Sardinha 2018). This includes blogs, micro-blogs, workplace emails, discussion posts, reader feedback, and online newspaper columns. Such work shows that web data is actually composed of several more or less distinct sub-registers, each specific to a given communicative context. This present paper approaches web data as a macro-register for the purpose of contextualizing the main registers of interest. Another recent line of work leverages the multi-dimensional approach to analyze and compare two Czech corpora: a carefully designed corpus and an opportunistic web-crawled corpus (Cvrček et al. 2020). The results show that traditional corpora provide a wider range of registers than web-crawled corpora, a somewhat different finding from other work on the complex registers found in web data (Egbert et al. 2015). This difference raises questions about how well findings from English generalize across languages.

The goal of this present paper is to extend our understanding of register to a fully multilingual setting. This involves quantifying relationships both within register-specific corpora (homogeneity) and between register-specific corpora (similarity) for 60 languages, with the same registers represented for each language using comparable corpora. Corpus similarity measures are best suited for this task, both because of our ability to make ground-truth predictions as validation and because we know that the specific features involved in register variation are unique to each language. For practical reasons, a manual interpretation of features is not a part of this paper.

## 3. Data and Methodology

The first task is to collect comparable corpora that represent each communicative situation. We call these sets of observations *register-specific corpora*: examples of usage from each context of production. The Wikipedia register (WK) is collected from the public Wikimedia dump of March 2020. The social media register (TW) is collected from Twitter using geo-referenced tweets. As discussed in Section 3.1, these two registers provide contrasting corpora that enable a cross-linguistic comparison. If register variation is universal, then the relationships between these corpora should remain stable across a diverse set of languages.

It is possible that geographic variation presents a confounding factor for a study like this. In other words, languages like English or French or Arabic are used in many different countries around the world. The comparison of registers would be distorted if the social media corpus represented British English but the web corpus used for contextualization represented American English. We control for geographic variation by deriving the web (CC) and TW corpora from a single country for each language. We choose the country which has the most data for that language. This geographic selection is shown for each language in Table 3. For example, the Amharic data is constrained to Ethiopia and the Somali data is constrained to Somalia. This geographic constraint controls for the possibility of geographic variation distorting our observation of register. Previous work has also taken into account native language for register (Kouwenhoven et al. 2018), a distinction that is not possible in this context.

While the main focus of this paper is on the TW and WK corpora, we need additional data from each language in order to compare those corpora. First, we compile a corpus of independent registers for each language for the purpose of selecting features. These corpora contain movie subtitles, news commentary articles, and Bible translations (Tiedemann 2012; Christodoulopoulos and Steedman 2015). These background corpora allow us to compare the relationship between TW and WK without deriving our representations from the same corpora being compared. An additional corpus representing a macro-web register (CC) is collected from the *Corpus of Global Language Use* (Dunn 2020), ultimately derived from the Common Crawl. The purpose of the web corpus is to contextualize the relationship between TW and WK across languages, providing a point of comparison. Thus, each language is represented by four comparable corpora: two register-specific corpora (TW, WK) for comparing register relationships, one background corpus for validation and feature selection, and one web corpus for contextualization.

## 3.1. Communicative Situations

We start by establishing a communicative profile for each of the two registers of interest. A break-down of different situational characteristics is given in Table 2. Despite some points of difference, these two situations are comparable across many situational parameters. For example, these are both written and digital sources of language use. Revisions are at least possible in each case, although less likely for social media and more likely for Wikipedia. Given the data collection methods, both contexts are public communications with an indefinite readership, although social media is more tailored for specific followers. The topics covered in each situation are also varied.

**Table 2. Communicative Situation For Registers**

|  | **TW** | **WK** |
|---|---|---|
| *Authors* | Single known author | Indefinite contributors |
| *Readers* | Friends, Followers | Seeking Information |
| *Channel* | Written, Digital | Written, Digital |
| *Production circumstances* | Revisions Possible | Revised and Edited |
| *Setting* | Public | Public |
| *Communicative purposes* | Multiple | Information |
| *Topic* | Not Limited | Not Limited |

For other parameters, however, these situations are distinct. For example, the author of a tweet is a single known individual (or at least a representative of that individual). But a Wikipedia article is drawn from potentially many contributors. The communicative purpose of Wikipedia is the most fixed: to present information in an objective manner. Twitter has a small number of purposes: to communicate with friends, to announce new pieces of information, to argue with or denounce other users. While both contexts cover many topics, Wikipedia has a goal of encyclopedic coverage so that it likely contains a broader range of topics.

Our point of comparison is a macro-register representing web pages, for which the communicative purpose is impossible to determine: a forum is a written conversation, a news article provides new information, a sales page markets some goods or services, a government website might provide basic information about policies and procedures. Even within a single sub-register drawn from web pages, the purpose is more variable than in other registers. Manual annotations of communicative purpose have shown that there is significant variation even within sub-registers that are very narrowly defined (Biber et al. 2020). By putting forward a single macro-register from web pages, we are unable to precisely define the communicative purpose. Register is a continuum and the boundaries that we draw to define them are somewhat arbitrary. Thus, we can imagine the web corpus as being a more generic macro-register that includes several sub-registers. This heterogeneity makes the web corpus an ideal point of comparison for situating TW and WK.

## 3.2. Languages and Features

Before we experiment with relationships within and between registers, we first evaluate the accuracy of the corpus similarity measures themselves. Part of this evaluation involves feature selection from independent corpora; this ensures that the measures do not over-fit the registers involved in the study, which would result in unstable features. The list of languages used is shown in Table 3, along with each language's family, script type, and morphological type. These three classifications for each language are included because these are all factors that may influence the performance of corpus similarity measures and the relationship between registers. Because Indo-European is a well-represented family, it is divided into branches (for example, *IE: Germanic*). The number of words across all corpora for each language is shown in the final column.

Languages are divided into four types of writing system: *Alphabetic* scripts use characters to represent individual phonemes. *Abjad* scripts use characters to represent consonants and leave vowels unrepresented. *Abugida* scripts represent consonant-vowel sequences together. Finally, *logographic* or syllabic scripts use characters that represent an entire word, morpheme, or syllable.

A broad morphological categorization for each language is also included in Table 3. There are four categories: *Agglutinative* languages have a range of different morphemes that retain the same form. *Fusional* languages combine multiple functions into a single morpheme. *Analytic* languages tend to use grammatical words instead of morphemes. Finally, we use the term *root-and-pattern* to describe Arabic and Amharic, which do not fit into the previous typology. Each language, of course, is a better or worse example of a particular type of morphology; the basic idea is to show that the experiments in this paper represent a diverse group of languages.

**Table 3. List of Languages by Family, Writing System, and Morphological Classification**

| Name | ISO | Family | Script | Morphology | Country | Words |
|---|---|---|---|---|---|---|
| Amharic | amh | Afro-Asiatic | Abugida | Root-Pattern | Ethiopia | 13,083,442 |
| Arabic | ara | Afro-Asiatic | Abjad | Root-Pattern | UAE | 38,148,360 |
| Vietnamese | vie | Austroasiatic | Alphabet | Analytic | Viet Nam | 44,831,262 |
| Indonesian | ind | Austronesian | Alphabet | Agglutinative | Indonesia | 32,748,733 |
| Malagasy | mlg | Austronesian | Alphabet | Agglutinative | Madagascar | 17,701,857 |
| Tagalog | tgl | Austronesian | Alphabet | Agglutinative | Philippines | 22,627,461 |
| Haitian | hat | Creole, French | Alphabet | Analytic | Haiti | 11,632,787 |
| Somali | som | Cushitic | Alphabet | Agglutinative | Somalia | 6,381,434 |
| Kannada | kan | Dravidian | Abugida | Agglutinative | India | 13,255,518 |
| Malayalam | mal | Dravidian | Abugida | Agglutinative | India | 16,794,478 |
| Tamil | tam | Dravidian | Abugida | Agglutinative | India | 29,216,255 |
| Telugu | tel | Dravidian | Abugida | Agglutinative | India | 12,041,648 |
| Albanian | sqi | IE:Albanian | Alphabet | Agglutinative | Albania | 33,582,132 |
| Bulgarian | bul | IE:Balto-Slavic | Alphabet | Fusional | Bulgaria | 37,809,641 |
| Czech | ces | IE:Balto-Slavic | Alphabet | Fusional | Czechia | 37,129,547 |
| Latvian | lav | IE:Balto-Slavic | Alphabet | Fusional | Latvia | 23,580,134 |
| Lithuanian | lit | IE:Balto-Slavic | Alphabet | Fusional | Lithuania | 28,777,596 |
| Macedonian | mkd | IE:Balto-Slavic | Alphabet | Fusional | N. Macedonia | 32,239,612 |
| Polish | pol | IE:Balto-Slavic | Alphabet | Fusional | Poland | 36,766,189 |
| Russian | rus | IE:Balto-Slavic | Alphabet | Fusional | Russia | 45,335,724 |
| Slovak | slk | IE:Balto-Slavic | Alphabet | Fusional | Czechia | 28,507,702 |
| Slovenian | slv | IE:Balto-Slavic | Alphabet | Fusional | Slovenia | 32,022,625 |
| Ukrainian | ukr | IE:Balto-Slavic | Alphabet | Fusional | Ukraine | 29,805,647 |
| Irish | gle | IE:Celtic | Alphabet | Fusional | Ireland | 9,262,764 |

| Danish | dan | IE:Germanic | Alphabet | Analytic | Denmark | 35,024,228 |
|---|---|---|---|---|---|---|
| German | deu | IE:Germanic | Alphabet | Fusional | Germany | 43,105,599 |
| English | eng | IE:Germanic | Alphabet | Analytic | United States | 61,075,309 |
| Icelandic | isl | IE:Germanic | Alphabet | Fusional | Iceland | 30,471,618 |
| Dutch | nld | IE:Germanic | Alphabet | Analytic | Netherlands | 39,172,924 |
| Norwegian | nor | IE:Germanic | Alphabet | Analytic | Norway | 32,525,826 |
| Swedish | swe | IE:Germanic | Alphabet | Analytic | Sweden | 34,842,984 |
| Greek | ell | IE:Hellenic | Alphabet | Fusional | Greece | 39,060,476 |
| Bengali | ben | IE:Indo-Iranian | Abugida | Fusional | India | 19,896,848 |
| Farsi | fas | IE:Indo-Iranian | Abjad | Analytic | Iran | 35,583,394 |
| Gujarati | guj | IE:Indo-Iranian | Abugida | Agglutinative | India | 13,885,950 |
| Hindi | hin | IE:Indo-Iranian | Abugida | Fusional | India | 34,521,582 |
| Marathi | mar | IE:Indo-Iranian | Abugida | Fusional | India | 14,305,246 |
| Punjabi | pan | IE:Indo-Iranian | Abugida | Fusional | India | 15,783,910 |
| Sinhala | sin | IE:Indo-Iranian | Abugida | Fusional | Sri Lanka | 19,919,824 |
| Urdu | urd | IE:Indo-Iranian | Abjad | Fusional | Pakistan | 23,952,579 |
| Catalan | cat | IE:Romance | Alphabet | Fusional | Spain | 20,568,207 |
| French | fra | IE:Romance | Alphabet | Fusional | France | 51,190,531 |
| Galician | glg | IE:Romance | Alphabet | Fusional | Spain | 10,102,290 |
| Italian | ita | IE:Romance | Alphabet | Fusional | Italy | 45,322,900 |
| Portuguese | por | IE:Romance | Alphabet | Fusional | Brazil | 43,677,195 |
| Romanian | ron | IE:Romance | Alphabet | Fusional | Romania | 38,796,683 |
| Spanish | spa | IE:Romance | Alphabet | Fusional | Colombia | 50,377,829 |
| Basque | eus | Isolate | Alphabet | Agglutinative | Spain | 11,191,128 |
| Japanese | jpn | Isolate | Logographic | Agglutinative | Japan | 36,616,359 |
| Korean | kor | Isolate | Logographic | Agglutinative | South Korea | 42,823,263 |
| Georgian | kat | Kartvelian | Alphabet | Agglutinative | Georgia | 18,575,314 |
| Mongolian | mon | Mongolic | Alphabet | Agglutinative | Mongolia | 15,458,196 |
| Chinese | zho | Sino-Tibetan | Logographic | Analytic | China | 17,117,201 |

| Thai | tha | Tai-Kadai | Abugida | Analytic | United States | 25,902,685 |
|---|---|---|---|---|---|---|
| Azerbaijani | aze | Turkic | Alphabet | Agglutinative | Azerbaijan | 15,733,448 |
| Turkish | tur | Turkic | Alphabet | Agglutinative | Turkey | 33,750,669 |
| Uzbek | uzb | Turkic | Alphabet | Agglutinative | Kazakhstan | 9,712,984 |
| Estonian | est | Uralic | Alphabet | Fusional | Estonia | 30,746,191 |
| Finnish | fin | Uralic | Alphabet | Agglutinative | Finland | 27,424,494 |
| Hungarian | hun | Uralic | Alphabet | Agglutinative | Hungary | 33,689,576 |

Recent work on corpus similarity measures has shown that a frequency-based approach with 5k bag-of-words features and Spearman's *rho* performs well across many languages (Li and Dunn 2022). Some languages achieve higher accuracy with word-based features and some with character-based features. Taking this finding as a starting point, we evaluate the accuracy of different feature types using those same parameters in Appendix 1 to validate these measures cross-linguistically. A frequency-based approach to corpus similarity uses a vector of frequency values, where each dimension in the vector represents a fixed vocabulary item. The Spearman *rho* has been shown to be highly accurate in comparing these frequency vectors, with similar corpora having a higher correlation coefficient. Unlike the chi-square (Kilgarriff 2001), this measure is not dependent on corpus size. Following previous work, we retain a fixed feature space for each language; this means that a vocabulary item is present in the vector for each corpus being compared, whether or not that item is observed in that particular corpus. Feature selection is frequency-based, with the most common 5k features being used (cf., Fothergill et al. 2016). For increased validity, we use the independent background corpora for feature selection to prevent overfitting. The details of this accuracy-based validation are found in Appendix 1.

For many languages, the best measure of corpus similarity relies on character n-grams. Table 4 provides an example of this kind of feature for English. Each column shows a different frequency strata: those features which would have been selected using only the top 100, 1k, 2k, and so on. The first column, the most frequent character n-grams, clearly captures function word information. As we move into less frequent strata, the features become grammaticalized words, such as *lot*, *either*, *trying*. Finally, we end in morph-like features like *civi* and *gned* which represent a number of different words (e.g., *civilized, civilization, civics*). The purpose of this table is to show that, although character-based features differ from traditional lexico-grammatical features, they nevertheless capture that same information when taken in the aggregate.

**Table 4. Example Features for English by Frequency Strata**

| 100 | 1k | 2k | 3k | 4k | 5k |
|------|------|------|------|------|------|
| the | lot | nced | cir | mers | role |
| ing | dia | eith | itut | dig | nfli |
| to | tati | tryi | rabl | intr | gned |
| you | resp | lia | unce | uspe | nifi |
| and | inv | ison | oto | rgen | emer |
| of | sor | sun | nden | civi | mann |

Corpus similarity requires comparing two sets of data. Previous work has shown that corpus similarity measures work robustly with sample sizes as small as 10k words; thus, we work with sub-corpora containing 10k words. We then create 250 unique pairs for each condition in our experiments, where each condition is a combination of registers. The advantage of creating 250 pairs of sub-corpora for each condition is that this allows us to measure the robustness or stability of a particular relationship. In other words, we observe a population of sub-corpora which allows us to better estimate the overall similarity or homogeneity of a corpus.

The distribution of the corpus similarity measures is language-specific; in other words, some languages have a high mean value and others a low mean value. Therefore we use the z-score to standardize the measure. A z-score is measured in standard deviation units, with positive z-scores indicating that the raw values lie above the mean and negative z-scores indicating that the raw values fall below the mean. Here the z-score is calculated for each language using pairs of sub-corpora from TW, WK and CC; the benchmark pairs here include three same-register conditions and three cross-register conditions. While our main focus is on the relationship within and between TW and WK, the CC corpus allows us to capture a wider population of corpora for the purposes of calculating the z-score. The result is a measure that is directly comparable across languages because it standardizes across the distribution of similarity values within each language.

### 3.3. Using Corpus Similarity to Analyze the Brown Corpus

Although the experiments in Appendix 1 show that the corpus similarity measures are robustly accurate across languages, how do they compare with factoring methods like multi-dimensional analysis? The basic distinction is that factor analysis performs dimension reduction in order to group related features together while corpus similarity measures compare samples instead of features. For example, corpus similarity measures could be used to cluster related sub-corpora together to form larger groupings. We undertake an example analysis with the Brown corpus (Kučera and Francis 1967) for the purpose of illustration (cf., Nini 2019).

**Figure 1. Similarities and Clusters for the Brown Corpus**

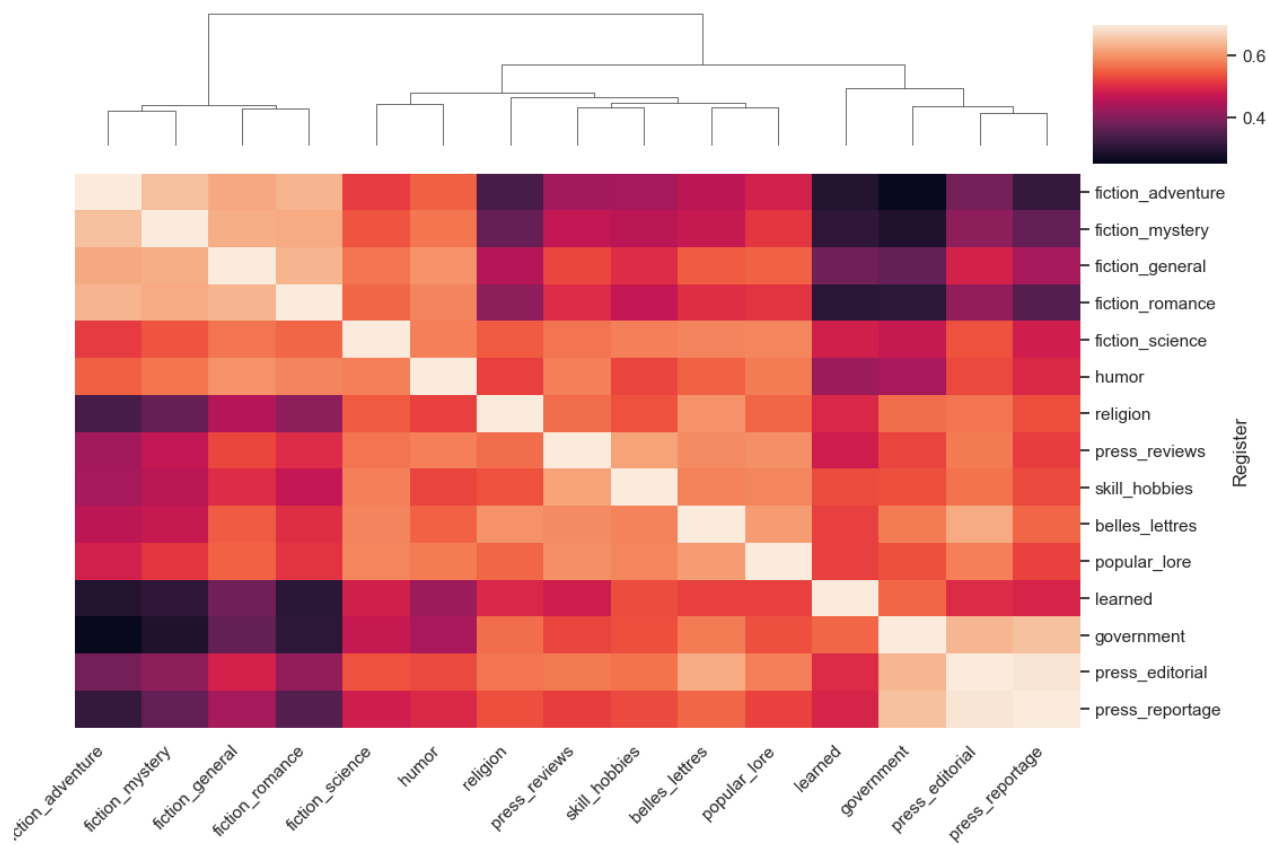

We start by using the corpus similarity measures evaluated in Appendix 1 to calculate the pairwise similarity between each of the 15 sub-sections of the Brown corpus (with 10k word sample sizes). No labels are used for this comparison; it is a purely-bottom up approach. These pairwise similarities are then clustered into groups using the Ward method for hierarchical clustering, producing the relationships shown in Figure 1. Each column and row represent the same sub-sections, aligned so that the diagonal represents self-comparisons. The first cluster makes a distinction between fiction and non-fiction registers. Within non-fiction, the second cluster makes a distinction between narrative non-fiction (like humor) and expository non-fiction (like news articles). While factor analysis would find the bundles of features which distinguish between these registers, corpus similarity measures robustly identify the aggregate relationships between the corpora themselves. In other words, given a network of pairwise similarity relationships between sub-corpora, a corpus similarity approach is able to cluster together functionally-related text types.

The purpose of analyzing the Brown corpus is to show that the underlying measures are not only robustly accurate across languages but are also capable of creating meaningful divisions between text types within a single corpus. These text types represent different communicative situations. In other words, the methods used here are capable of capturing increasingly fine distinctions between register-specific corpora in a way that builds groups of registers in a bottom-up manner. The main advantage over factor analysis is that this approach can be scaled and validated in a highly multilingual setting.

## 4. Analysis

This section experiments with register variation across 60 languages. We start by evaluating the homogeneity of TW and WK using the CC corpora as a reference point (Section 4.1). The idea is that corpora representing the same communicative situation should be equally homogeneous across languages. We then situate TW and WK within a two-dimensional space, again using CC as a reference point (Section 4.2). The idea is that, given the stability of the relationship between these communicative situations across languages, we should also see stability in the relationship between register-specific corpora. This experiment will tell us whether the properties of individual registers are consistent across languages. Taken together, these experiments provide a register profile for each language that is directly comparable across all 60 languages.

## 4.1. Homogeneity of Register-Specific Corpora

This section analyzes the self-similarity of register-specific corpora that have been produced under the same situational parameters. In this context, *self-similarity* refers to the population of pairwise similarity values between subsets of the same register-specific corpus. We take a larger corpus and divide it into hundreds of random sub-corpora. We then measure the similarity between these sub-corpora. The underlying hypothesis is that more homogenous registers (like Wikipedia) should produce homogenous corpora. And registers that are in fact collections of varied sub-registers (like the web) should produce more heterogeneous corpora.

For each register-specific corpus, we extract 250 pairs of unique sub-corpora, where each sub-corpus contains 10k words. Each pair of sub-corpora is represented using the similarity measure that is calculated using the frequency-based methods described above. We then use the z-scores for each language to normalize these similarity values to make them directly comparable across languages. Higher z-scores (usually positive values) imply that the corresponding pairs are more similar. This is shown in Figure 2 for English as a categorical scatterplot. The y-axis represents similarity, with more similar corpora having values toward the top. Each point represents a single pair of sub-corpora. Each column represents a different comparison: blue represents TW and green WK. The orange and red columns represent a comparison between TW-WK and WK-TW; because these pairs are symmetrical, we expect these two columns to be comparable.

*Figure 2. Homogeneity for English*

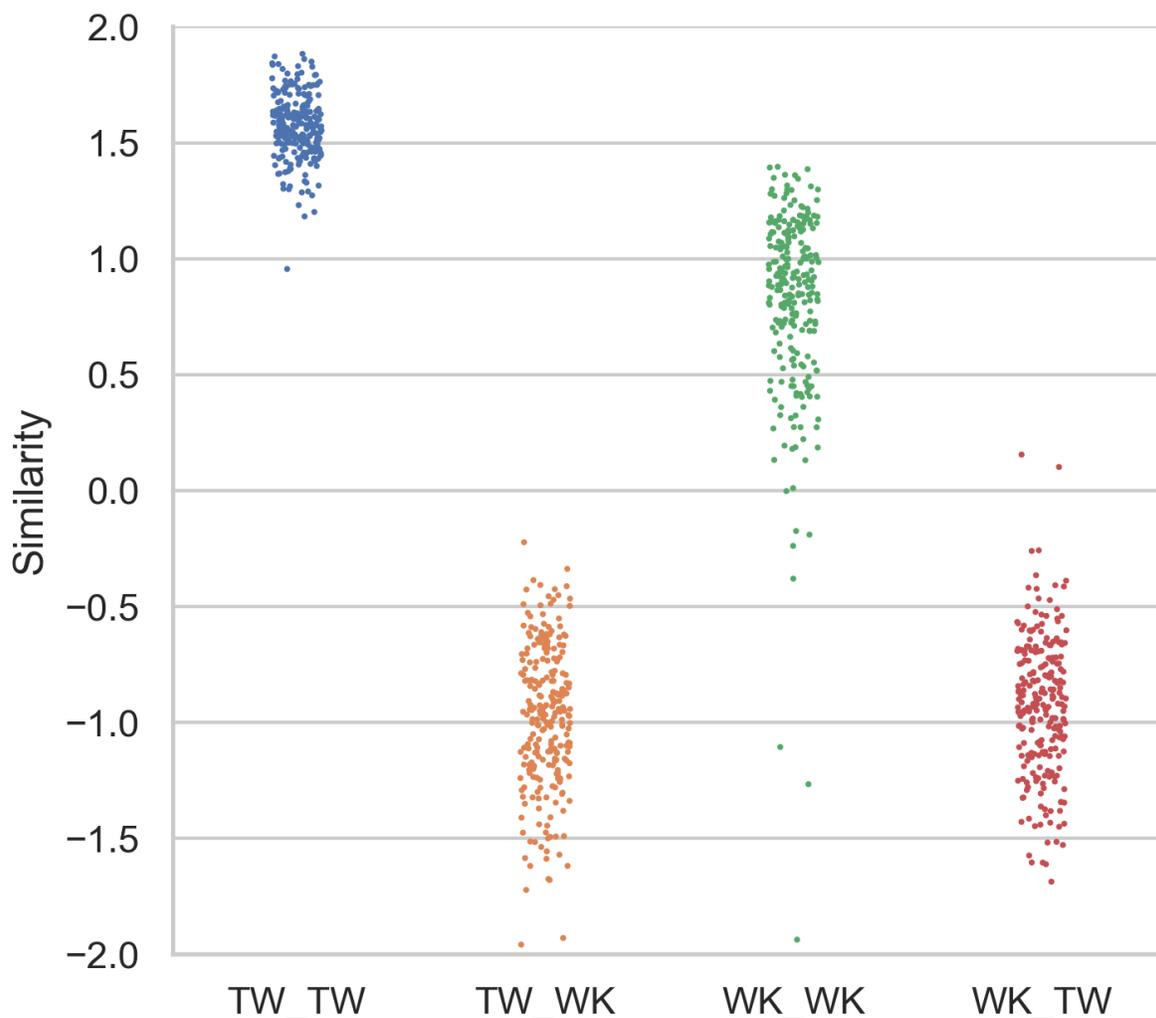

For English, TW is a highly homogenous corpus; this is true both in terms of its central tendency and in the limited number of outliers. On the other hand, WK also has a high central tendency, but there are also a small number of outliers which are quite dissimilar from other sub-corpora. This can be contrasted with the cross-register comparisons (TW-WK and WK-TW), which are equally dissimilar (thus, showing low similarity values). This figure also shows the advantage of a corpus similarity approach that allows us to make many measurements from a larger corpus; for example, a single representation of WK would disguise these outliers. To the degree that each communicative situation is homogenous cross-linguistically, we would expect that this same pattern of self-similarity would remain stable as well.

**Figure 3. Homogeneity for Arabic**

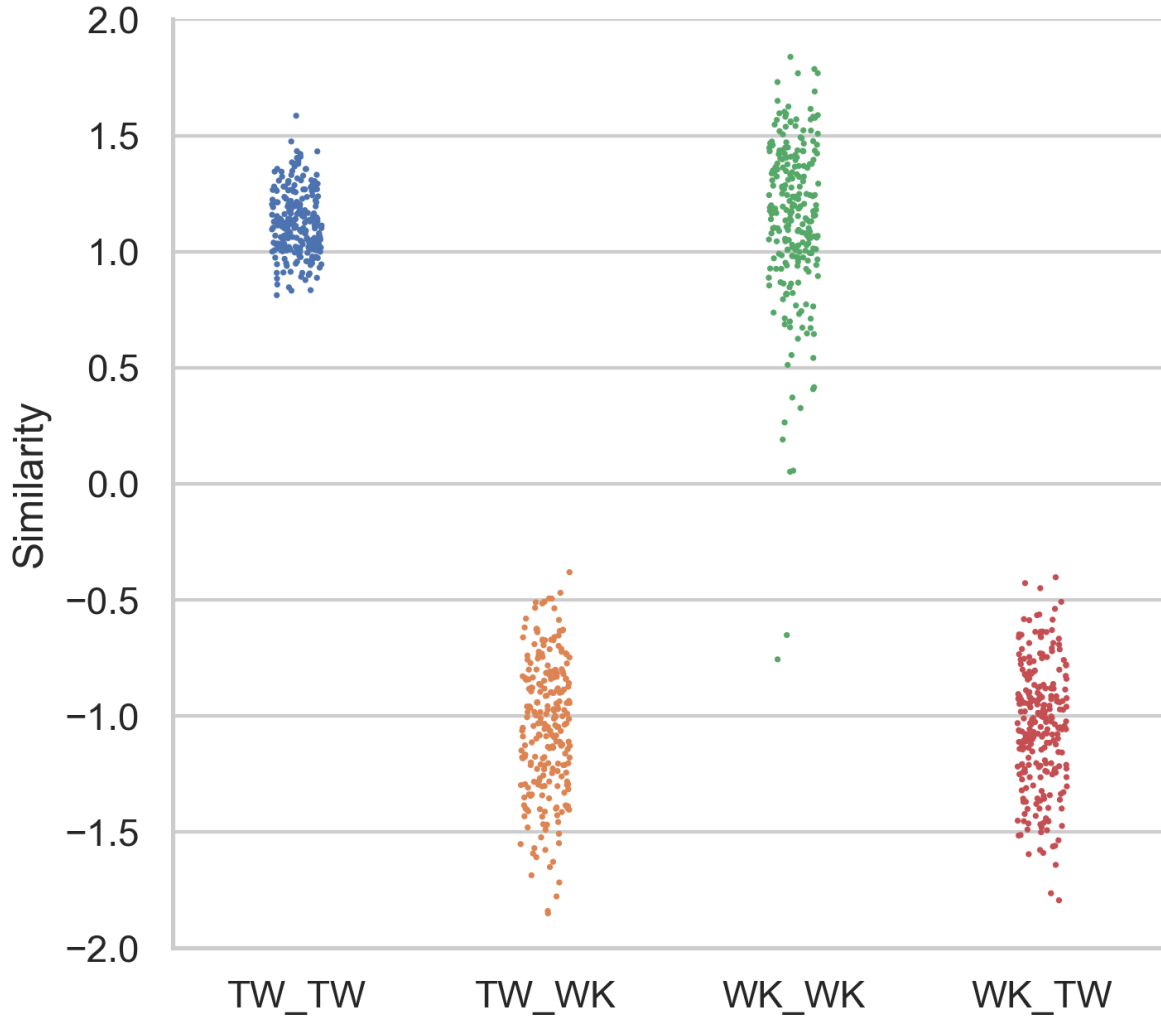

We next investigate Arabic, an unrelated language that differs significantly from English both in its writing system and its morphology. As shown in Figure 3, however, the same pattern remains: TW is the most homogenous, with few outliers, and WK has a similar central tendency but with more outliers (i.e., subsets of the corpus which are dissimilar to most of the corpus).

**Figure 4. Homogeneity for Indonesian**

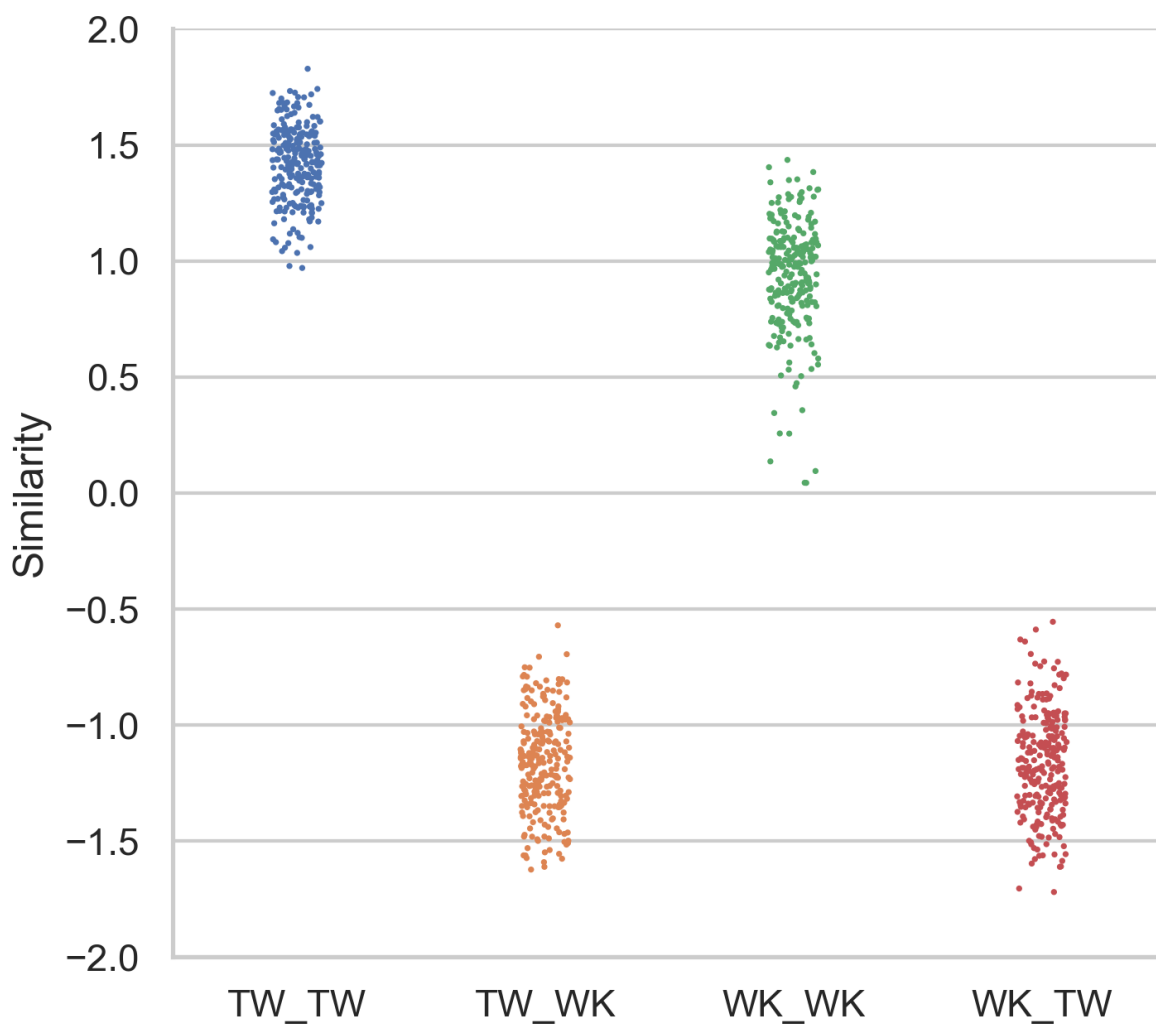

The same representation is shown in Figure 4 for Indonesian, a widely spoken Austronesian language which, however, is less commonly studied in corpus linguistics. Here we find a closely comparable pattern: TW is the most homogenous with few outliers while WK has a small number of outliers; the cross-register comparisons are clearly distinguished from the same-register comparisons. A final language, Korean, is shown in Figure 5, chosen as a representative of yet another type of script. Here the same pattern is shown, with however a small number of outliers within TW.

**Figure 5. Homogeneity in Korean**

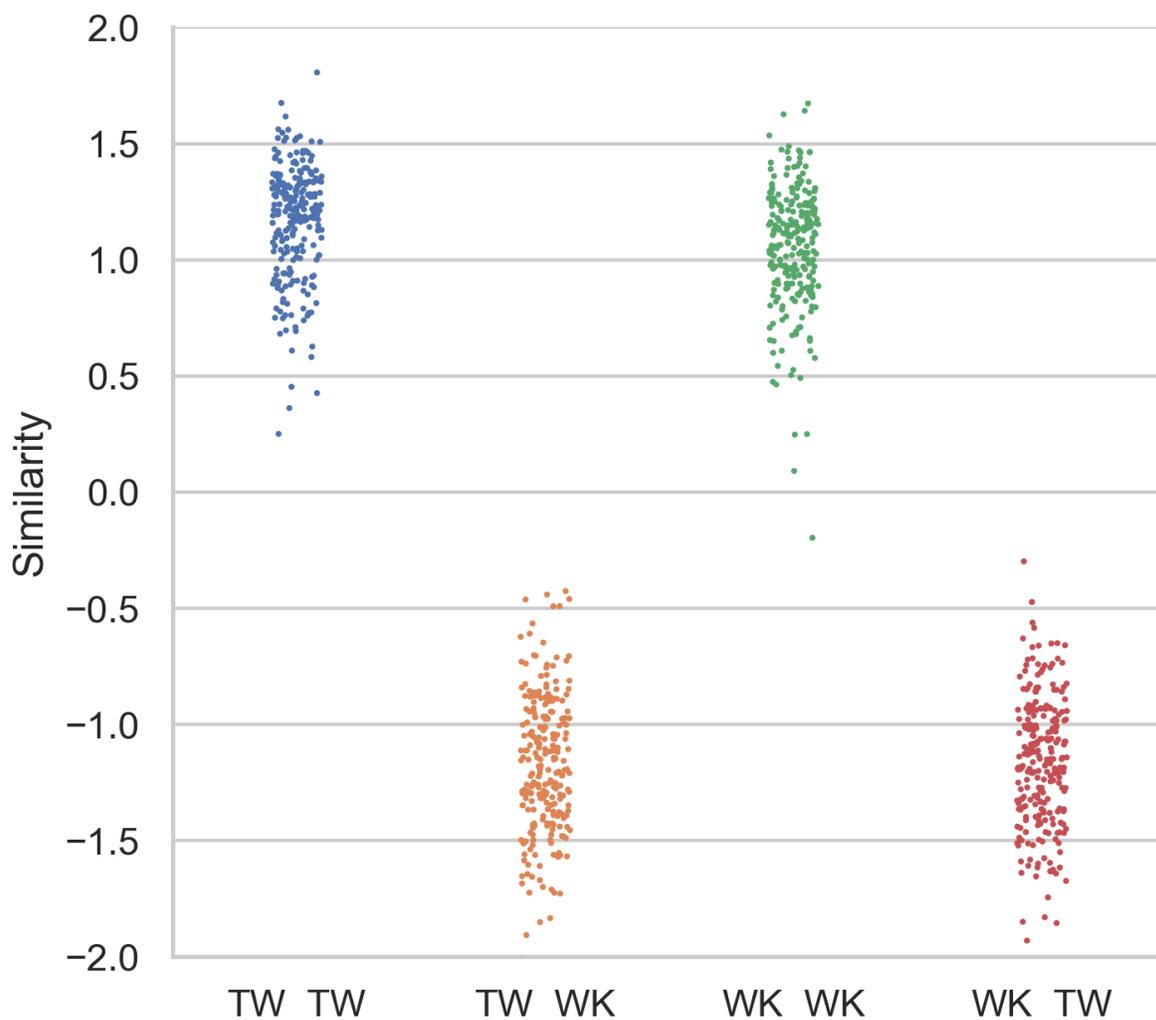

Given these patterns, we put forward a single metric for homogeneity that is based on the mean z-score similarity for a particular corpus, as shown for Figure 6. This figure contains CC as an additional comparison point, a macro-register which is hypothesized to contain more internal variation because it has been produced under a wider range of communicative situations. Here the mean is calculated across 100 pairs of sub-corpora using a Bayesian approach with a 90% confidence level. This allows us to control for cases where there is higher variation across pairs of samples. The higher this score, the more homogenous the corpus is. In addition to the multi-lingual plot in Figure 6, the complete set of language-specific figures is available in the supplementary material.

**Figure 6. Homogeneity by Corpus for 60 Languages**

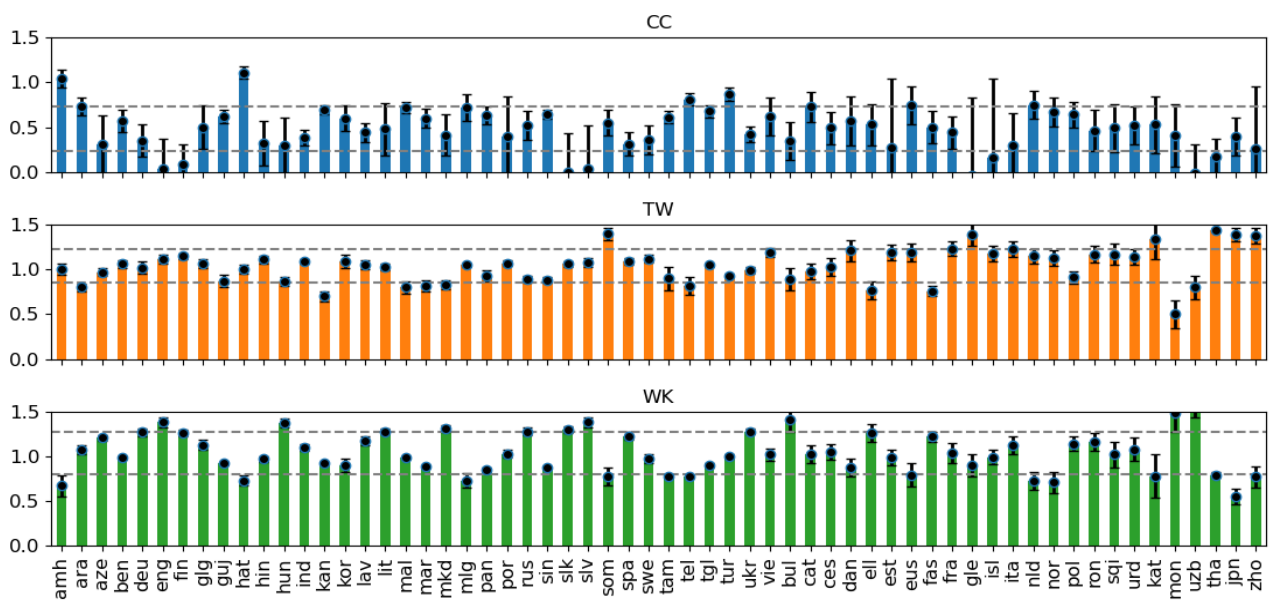

Because we are interested in a cross-linguistic approach to register, we visualize the spread of homogeneity for each register-specific corpus across all 60 languages in Figure 6. Here each register-specific corpus is a separate subplot. The mean similarity is used to represent homogeneity, with the standard deviation shown with error bars. The dotted lines indicate those values falling within one standard deviation of the mean across all languages.

We see that, for TW and WK, most languages fall within one standard deviation of the mean. In other words, the registers that these corpora represent are equally homogenous across languages. The web corpus, however, is the most heterogeneous, showing both lower self-similarity scores and a wider variation across languages. These results are what we expect given that this corpus in fact represents more than one register (Biber et al. 2020): there is more variation within this corpus because it represents a family of sub-registers in a way that the other corpora do not.

This section has shown that there is a stable relationship within registers when viewed using a standardized corpus similarity measure. The diverse range of languages observed provides clear evidence that the expected patterns of register variation are universal. In other words, comparable corpora drawn from the same communicative situations have the same relationships across languages even though the features which carry those relationships are unique to each language.

## 4.2. Relationships between Register-Specific Corpora

In this section we focus on relationships between corpora drawn from different communicative situations, going beyond distance to include the direction of distance as well. The question is whether the stable relationship between these register-specific corpora is driven by the underlying communicative situation. If so, this indicates that register variation remains universal across languages. As before, we estimate the relationship between two corpora using 250 pairs of unique sub-corpora.

This is shown in Figure 7, for English, as a scatterplot. Each point represents a single pair of sub-corpora, allowing us to visualize not only the general distribution but also the potential for outliers. For each point, the y-axis represents the standardized distance between that sample and a random sample from TW; thus, for TW (in blue) this represents homogeneity. Similarly, the x-axis represents the standardized distance between that sample and a random sample from WK; thus for WK (in orange) this represents homogeneity. The web corpus is provided as a reference point; given the variation in situational parameters behind the web corpus, we expect that the pattern of CC (in green) is less stable across languages than the patterns for TW and WK.

**Figure 7. Register Relationships in English**

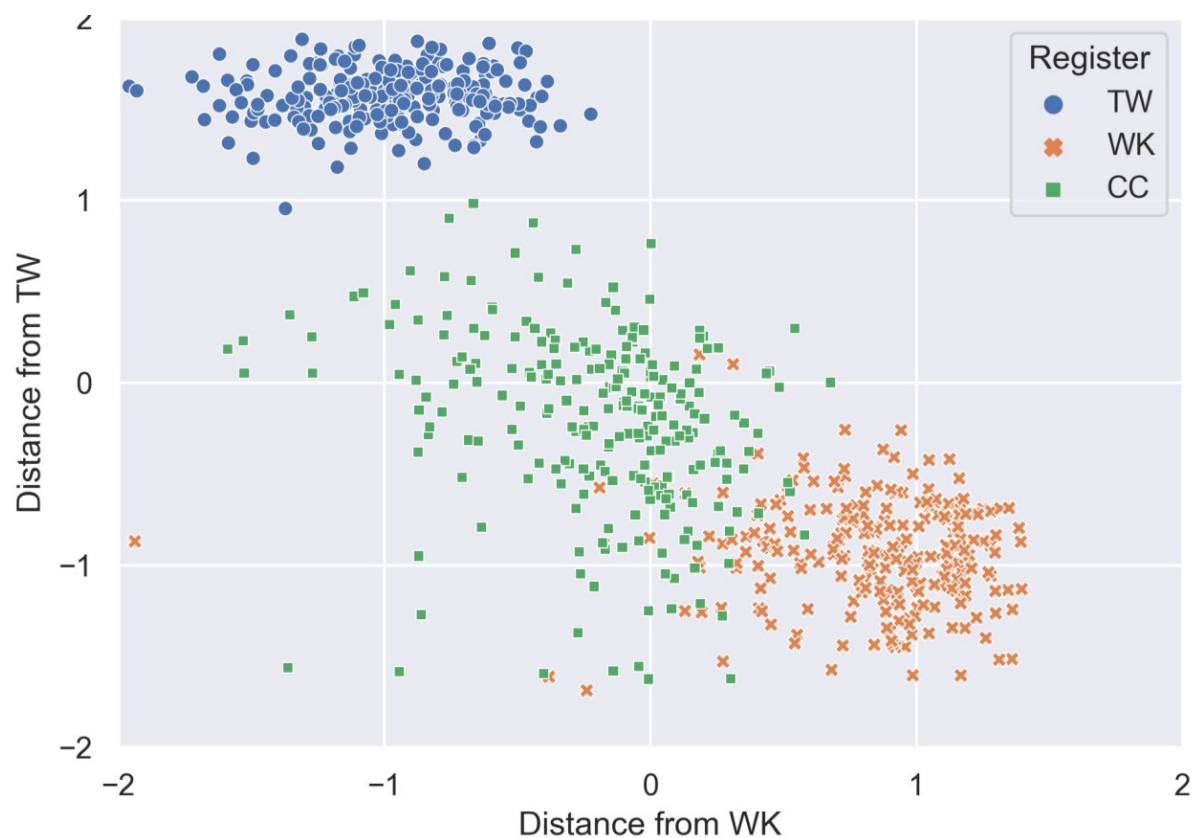

For English, in Figure 7, there is a clear separation between TW and WK. The web corpus, however, intermingles with the WK corpus and, in the case of some outliers, extends to the boundaries of TW. This is a more detailed reflection of the heterogeneity of the web corpus (cf., Figure 6) but here expanded to cover two-dimensions. As before, TW remains the most homogenous register, even given this more detailed view, with WK showing a number of outliers. These two dimensions of comparison, of course, represent relationships between samples rather than relationships between specific linguistic features.

A different profile is shown by Amharic in Figure 8. Here the relationship between TW and WK remains comparable with English, but this time the web corpus intermingles with TW rather than with WK. Thus, those corpora which represent stable communicative situations maintain a consistent relationship; but the web corpus, as a macro-register, is here more similar to TW.

**Figure 8. Register Relationships in Amharic**

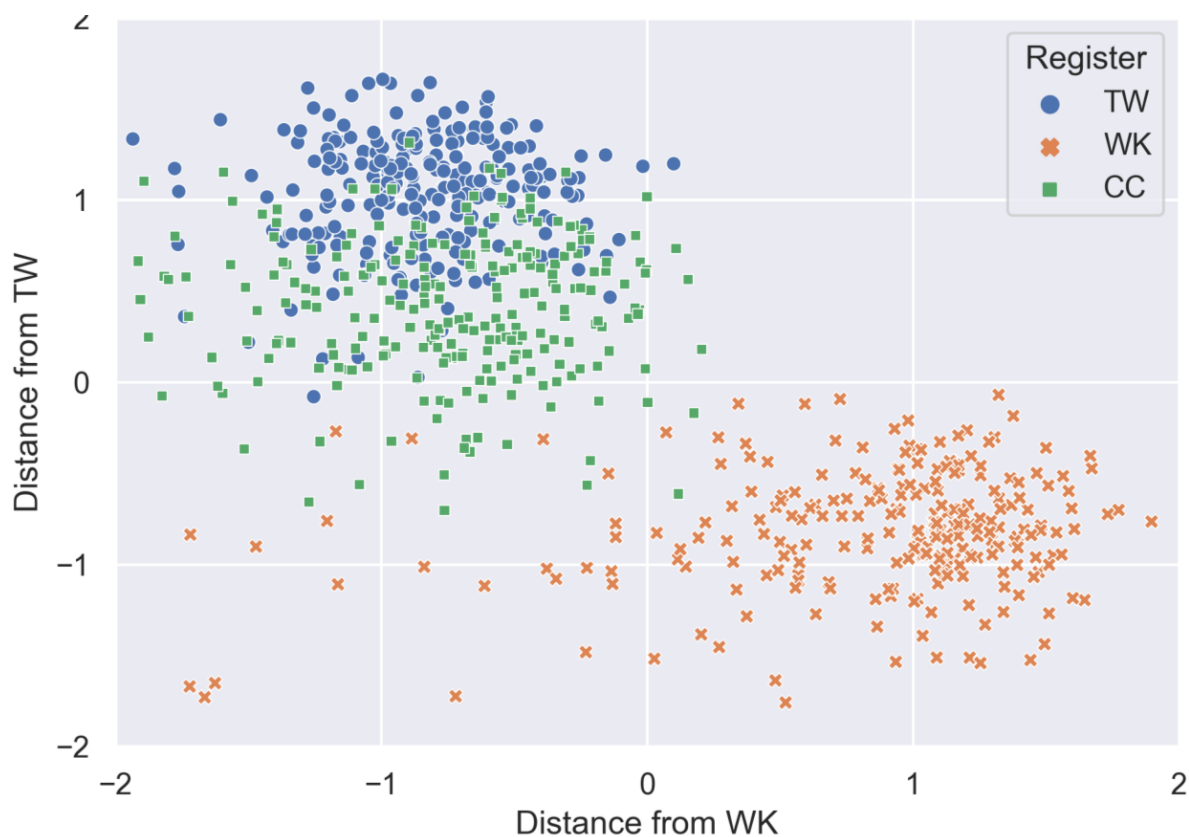

A third configuration is shown in Figure 9 for Tagalog. In this case, TW and WK remain as always clearly separated. Here, however, the web corpus also forms a discrete distribution, evenly situated in the center with relatively few outliers compared to other languages. This suggests that, in Tagalog as in all other languages, the communicative situation behind TW and WK produces similar linguistic relationships between these corpora. However CC is also a homogeneous corpus indicating that the web register in Tagalog is capturing a smaller range of communicative situations than it is in other languages.

**Figure 9. Register Relations in Tagalog**

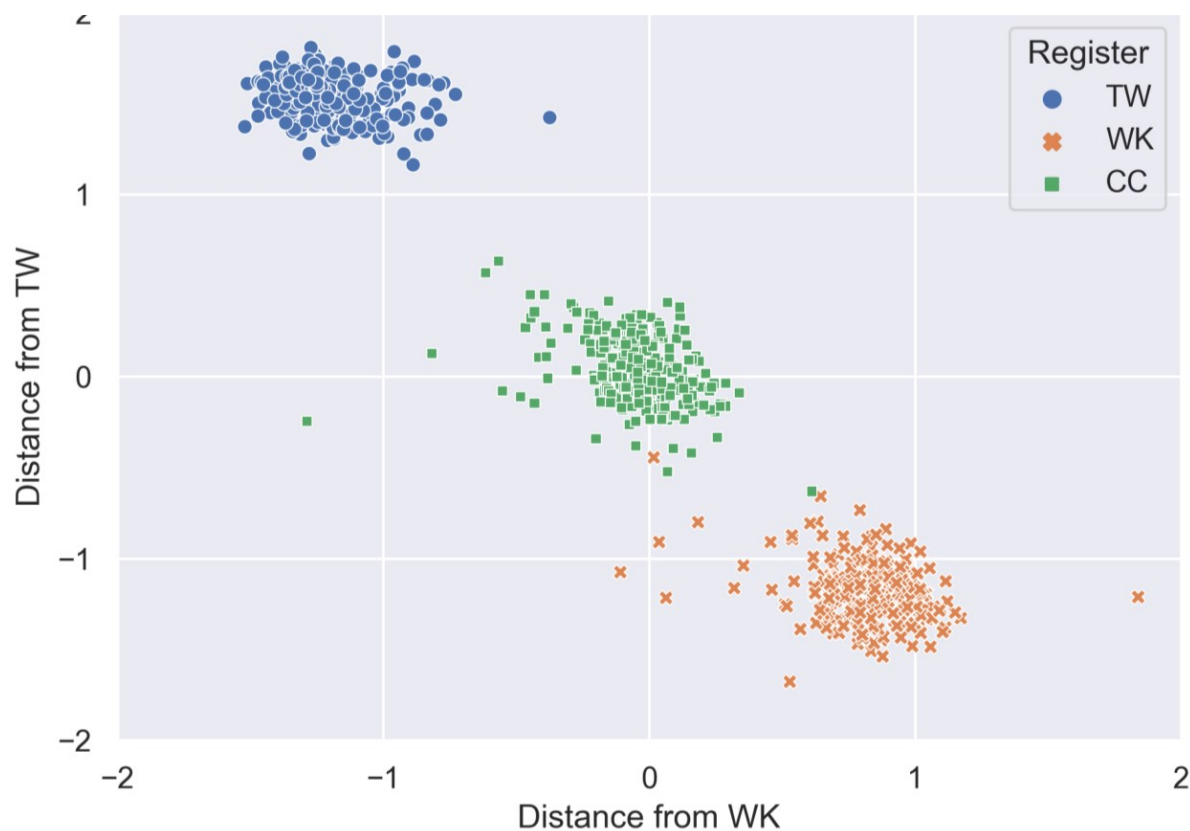

A final pattern that is shown across languages is displayed by Hindi in Figure 10. Here the distribution of the web corpus (in green again) is so closely intertwined with WK that there is no clear separation between the two. We might expect, for example, that the accuracy in distinguishing CC and WK in Hindi would be lower. At any rate, there is a clear relationship between CC and WK here. Interestingly, this same configuration is shown by a number of languages from South Asia: Bengali, Gujarati, Kannada, Malayalam, Marathi, Punjabi, Sinhala, Tamil, Telugu. These languages represent two diverse language families (Indo-Iranian and Dravidian), but include all of the languages drawn geographically from India. A closely related Indo-Iranian language, Urdu, is instead drawn from Pakistan and does not show this same configuration. While the relationship between TW and WK remains stable here, there is an interesting geographic and non-genetic pattern in the distribution of web corpus.

**Figure 10. Register Relations in Hindi**

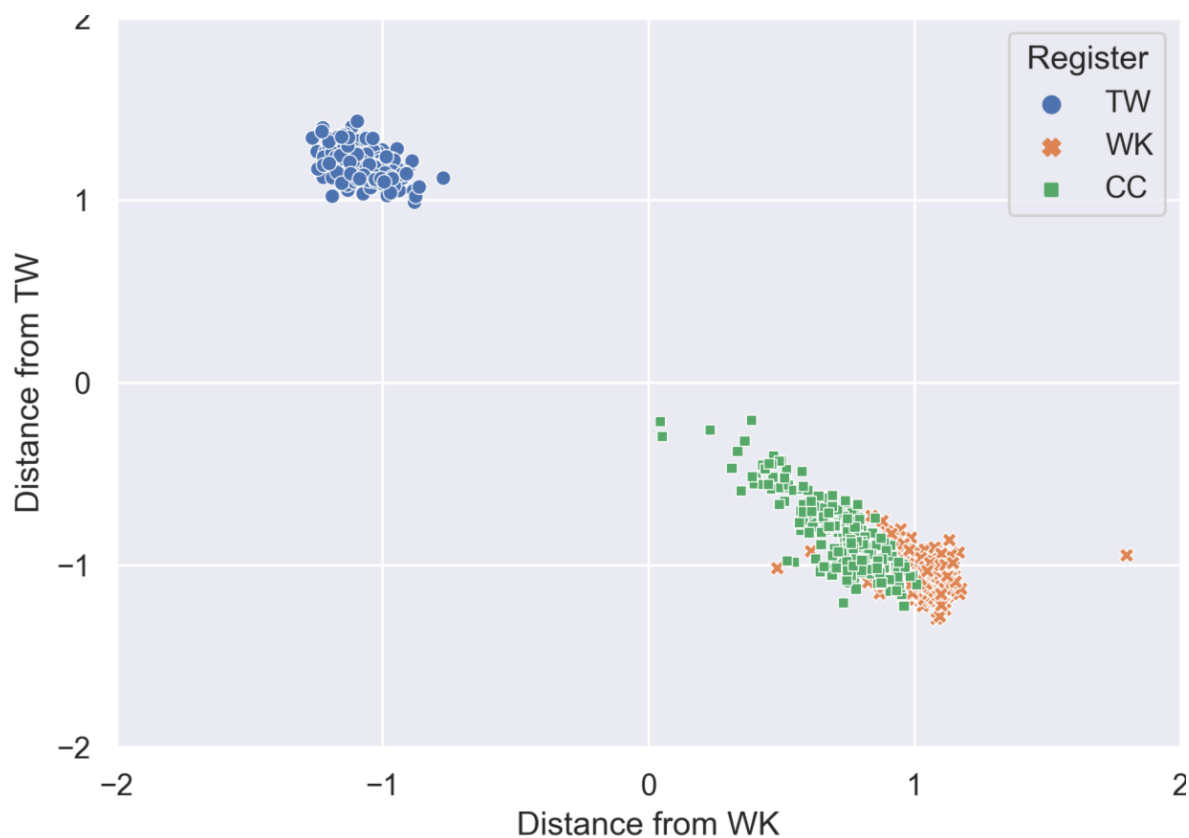

The analysis in this section has taken a closer look at the relationships between register-specific corpora from a cross-linguistic perspective. The two registers of interest, TW and WK, have the same situational parameters across languages. We thus expect that there will be the same linguistic relationships between them, even though the features involved are unique to each language. The figures for each language are available in the supplementary material. This consistency across languages confirms the hypothesis that register variation is universal.

## 5. Discussion

The experiments in this paper are designed to test the theoretical construct of register in a highly multilingual setting. The essential question is whether the impact of register is a core cross-linguistic property of language. Previous work has shown that register is a core property of a number of individual languages when viewed in isolation. We know that languages for which we have abundant corpora contain a large number of distinct varieties defined by their context of production: for instance, Czech (Cvrček et al. 2020), English (Egbert et al. 2015), and Portuguese (Sardinha et al. 2014). We also know that less-studied languages like Somali also contain distinct registers, thus showing the impact of register variation (Biber, 1995). Previous language-specific work, therefore, suggests that all languages have distinct registers. In other words, all languages have varieties that are defined by their context of production.

The experiments in this paper have gone a step further and shown that there is also a stable and predictable relationship between registers across languages. In other words, given the same sets of registers (language use collected from similar contexts of production), there is a predictable relationship between the resulting register-specific corpora. This is an important finding: register itself is a theoretical construct and, as such, cannot be directly observed. Our approach in this paper has been to take register-specific corpora as observations of registers so that corpus similarity measures (quantifying the relationships between corpora) can be used to estimate relationships between the underlying registers themselves. Since the comparable corpora used in this study represent the same contexts of production across 60 languages, individual languages (e.g., English, Arabic, Russian, etc) provide individual observations of the relationship between these registers.

First, if these registers exist as distinct varieties across all 60 languages then the accuracy for distinguishing between samples from each register-specific corpus should be relatively high across languages (cf., Appendix 1). This is, in fact, the case. Given previous findings from language-specific studies of register, this conclusion is not surprising; however, it does confirm our expectation that register is a stable and universal property of language.

Second, and more importantly, if the same register profile is stable across languages, it means that register as a latent variable leads to the same observable relationships between register-specific corpora. The homogeneity of register-specific corpora indicates the degree to which a register is subject to internal variation. The similarity between register-specific corpora indicates the degree to which registers are related to one another, including the discreteness of their boundaries.

Our first experiment focused on homogeneity as the self-similarity of a register-specific corpus. The situation behind each corpus remains the same across languages. For two of our corpora, social media (TW) and Wikipedia (WK), the underlying situation is relatively homogenous. For the web corpora (CC), however, we expect that there will in fact be a number of distinct sub-registers represented, ranging from sales prose to forums to question-answering pages. Previous work has shown that there is a continuous range of registers for web corpora, with an increasing number of closely related and difficult to distinguish sub-registers as the criteria for defining them become more specific (Biber et al. 2020). This property of the web corpora is captured in our estimates of homogeneity, which show us the degree to which our register-specific corpora actually represent a single register: TW and WK are internally consistent but CC is not. The goal in these experiments is not to determine the threshold at which a specific register needs to be divided into multiple labels. Instead we take a continuous approach, so that corpora which represent multiple contexts of production become more heterogeneous.

As shown in Figure 6, in all but two languages, the homogeneity of the web corpus is lower than the other two corpora; the two exceptions, Amharic and Haitian, have relatively small web corpora. As shown in Figure 7, there is a broader distribution of similarity values from the web corpora in some languages (like Amharic, *amh*, Figure 8) than others (like Tagalog, *tgl*, Figure 9). These figures thus show the degree to which each corpus actually does represent a single coherent context of production. These experiments show that the homogeneity or consistency of each register, as represented by register-specific corpora, is relatively stable across a large set of diverse languages.

For our second set of experiments, we calculate a register profile for each language as visualized in Figure 7. Again, the idea is to use relationships between observable register-specific corpora to estimate relationships between the underlying registers. The register profile presented in Figure 7 provides an indication of both (i) homogeneity, the distribution of each cluster as discussed above, and (ii) similarity, the distance between clusters. The context of production is the same for all 60 languages, because each set of corpora is comparable. Our hypothesis is that the same context of production leads to the same variety which should lead to the same relationships between observed register-specific corpora. As we observe register-specific corpora across a large number of languages, we would expect to find a random distribution of relationships if there was not a stable and predictable cross-linguistic phenomenon of register. As before, these results provide evidence for register as a universal property of language. Not only do all 60 languages have the same distinct registers here, but those registers produce corpora with predictable relationships.

The results in this paper make a significant contribution to linguistic theory by showing that the context of production has a systematic influence across languages, resulting in similar register profiles. Previous work has focused on a close examination of one or two languages in isolation. Such work motivates the idea of a connection between linguistic variants and the context of production, but depends quite heavily on English and is unable to establish the systematicity of register variation across languages.

The experiments in this paper show how far the basic pattern of register transfers across languages. This is an important finding because it provides further evidence that register variation is a fundamental and predictable attribute of language with a consistent influence on grammar and the lexicon. While previous work has established the importance of register within individual languages, the contribution of this paper to linguistic theory is to show how this extends across a diverse set of languages in a predictable and systematic manner.

**Appendix 1. Validating Corpus Similarity Measures**

This appendix describes validation experiments used to ensure that the corpus similarity measures provide robust measurements across the 60 languages discussed in the main paper. To evaluate the measures, we quantify the degree to which they make accurate predictions about the boundaries between corpora using a simple threshold. In other words, can corpus similarity measures be used to predict whether two sub-corpora come from the same or from different sources? This task (introduced by Kilgarriff 2001) provides a ground-truth validation for both the corpus similarity measures and the linguistic features they depend on.

The first step is to determine the best feature type for each language, using the independent background corpora described in the main paper for feature selection. We evaluate word 1-grams, word 2-grams, character 3-grams, and character 4-grams for each language. To ensure robustness, we employ a cross-validation framework: the corpora are divided into training and testing sets five times, until each subset of a corpus has appeared in the test set once. We average the accuracy of predictions across these five folds and choose the feature type for each language that achieves the highest accuracy.

The similarity measure based on Spearman's *rho* returns a continuous value. To convert this into an accuracy evaluation, we set a threshold for making predictions about whether two input samples come from the same corpus or from different corpora. The more often this threshold leads to correct predictions, the more accurate the measure is. In other words, we draw samples from three distinct corpora (TW, WK, CC). We then use the similarity measures, together with a threshold, to predict whether two samples came from the same corpus. Measures with a high prediction accuracy are able to distinguish between same-corpus and cross-corpus pairs. We draw on previous methods for estimating the optimum thresholds, methods which have been demonstrated to work well in related problems (Nanayakkara and Ranathunga 2018; Leban et al. 2016).

The threshold calculation is shown below. We take the lowest average similarity for same-register pairs (for example, maybe CC-CC is the least homogenous register). Then we take the highest average similarity for different-register pairs (for example, maybe CC-WK are the most similar registers). The threshold is set halfway between these minimum and maximum values. This threshold is calculated on the training data for each fold.

$$T = \frac{1}{2}(\min(Similarity_{CC-CC}, Similarity_{TW-TW}, Similarity_{WK-WK}) + \max(Similarity_{CC-WK}, Similarity_{TW-WK}, Similarity_{CC-TW}))$$

The main experiments in the paper do not require a threshold for calculating accuracy because we are concerned with continuous relationships within and between register-specific corpora. However, here we evaluate accuracy because this allows us to determine how meaningful these measures are for the underlying task. For example, if corpus similarity measures for Mongolian make poor predictions about register boundaries, this tells us that our measure is not suitable for the comparison of register-specific corpora in Mongolian. Thus, the accuracy evaluation based on cross-fold validation ensures the robustness of the experiments in the main paper. This provides a cross-linguistic ground-truth to support our analysis.

We start by verifying the accuracy of these corpus similarity measures using the cross-fold validation experiment described above. The results are shown in Table A, together with the best feature type for each language. The accuracy value here is the average accuracy across training-testing folds for the corresponding feature type: W1 represents word 1-grams, C2 represents character 2-grams, and so on. For some languages, there are more than one feature type that produces the same or similar accuracy. For example, Bulgarian has similar accuracies with both W1 and C4 (98% vs 97%) and Amharic has four types (C2, C3, C4, W1) that all achieve 100% accuracy. In the case of ties, we prefer *character* features over *word* features. In the case of a further tie, we prefer a higher n-gram (e.g., 4 over 3).

This selection procedure gives a single best measure for each language. The accuracies range from 88% (Japanese and Hindi) to 100% (among others, Amharic and Bengali). Overall, 49 of 60 languages achieve 95% accuracy or higher; and all languages are above 88% accuracy. When a language has lower accuracy, this means that the boundary between two of the registers is not distinct using a similarity measure. For example, if the CC and TW corpora are very similar, then some samples of each will be misidentified. This means that, for our purposes, an accuracy of 88% is not problematic, rather indicating that the relationship between registers in this language is not as distinct as in other languages.

**Table A. Accuracy and Best Feature Type by Language**

| Language | Features | Accuracy | Language | Features | Accuracy |
|----------|----------|----------|----------|----------|----------|
| amh | C4 | 100% | lit | C4 | 99% |
| ara | C4 | 99% | mal | C4 | 100% |
| aze | C4 | 96% | mar | C4 | 94% |
| ben | C4 | 100% | mkd | C4 | 99% |
| bul | W1 | 98% | mlg | C4 | 100% |
| cat | W1 | 100% | mon | W2 | 94% |
| ces | W1 | 98% | nld | W1 | 100% |
| dan | W1 | 99% | nor | W1 | 98% |
| deu | C4 | 98% | pan | C4 | 99% |
| ell | W1 | 97% | pol | W1 | 99% |
| eng | C4 | 98% | por | C4 | 98% |
| est | W1 | 98% | ron | W1 | 99% |
| eus | W1 | 100% | rus | C4 | 100% |
| fas | W1 | 96% | sin | C4 | 100% |
| fin | C4 | 94% | slk | C4 | 94% |
| fra | W1 | 100% | slv | C4 | 96% |
| gle | W1 | 90% | som | C4 | 100% |
| glg | C4 | 100% | spa | C4 | 99% |
| guj | C4 | 95% | sqi | W1 | 96% |
| hat | C4 | 100% | swe | C4 | 96% |
| hin | C4 | 88% | tam | C4 | 96% |
| hun | C4 | 95% | tel | C4 | 100% |
| ind | C4 | 99% | tgl | C4 | 100% |
| isl | W1 | 93% | tha | C3 | 90% |
| ita | W1 | 94% | tur | C4 | 100% |
| jpn | C2 | 88% | ukr | C4 | 99% |
| kan | C4 | 98% | urd | W1 | 100% |
| kat | W2 | 96% | uzb | W2 | 99% |
| kor | C4 | 99% | vie | C4 | 100% |
| lav | C4 | 99% | zho | C2 | 96% |

This accuracy-based evaluation tells us that the similarity measures make robust distinctions between register-specific corpora across all 60 languages, with some languages being 100% accurate and others retaining a small number of misclassifications. This prediction-based validation gives us confidence in the ability of these measures to capture variation within these languages.